\documentclass{article}
\usepackage{spconf,amsmath,graphicx}
\usepackage[font=small,skip=1pt]{caption}
\usepackage{comment}
\usepackage{amsmath}
\usepackage{amssymb}
\usepackage{caption}
\usepackage{subcaption}
\usepackage[linesnumbered,ruled,vlined]{algorithm2e}
\usepackage{comment,color,soul}
\usepackage{color, colortbl}
\usepackage{subcaption}
\usepackage{mathtools}
\usepackage{amsfonts}       %
\usepackage{amsmath}        %
\usepackage{mathtools}      %
\usepackage{multirow}
\usepackage{nicefrac}       %
\usepackage{microtype}

\title{AttentionLite: Towards\\ Efficient Self-attention models for Vision}
\name{Souvik Kundu$^{1}$$^{\ddagger}$ , Sairam Sundaresan$^{2}$\thanks{$^{\ddagger}$Work done during internship at Intel Labs.} \thanks{Arxiv pre-print. Manuscript under review.}}
\address{$^{1}$University of Southern California, Los Angeles, CA\\$^{2}$Intel Labs, San Diego, CA}
\begin{document}
%
\maketitle
\begin{abstract}
 We propose a novel framework for producing a class of parameter and compute efficient models called AttentionLite suitable for resource constrained applications. Prior work has primarily focused on optimizing models either via knowledge distillation or pruning. In addition to fusing these two mechanisms, our joint optimization framework also leverages recent advances in self-attention as a substitute for convolutions. We can simultaneously distill knowledge from a compute heavy teacher while also pruning the student model in a single pass of training thereby reducing training and fine tuning times considerably. We evaluate the merits of our proposed approach on the CIFAR-10, CIFAR-100 and Tiny-ImageNet datasets. Not only do our AttentionLite models significantly outperform their unoptimized counterparts in accuracy, we find that in some cases, that they perform almost as well as their compute-heavy teachers while consuming only a fraction of the parameters and FLOPs. Concretely, AttentionLite models can achieve up to $30\times$ parameter efficiency and $2\times$ computation efficiency with no significant accuracy drop compared to their teacher.          
\end{abstract}

\section{Introduction}
\label{sec:intro}
Convolutional neural networks (CNNs) have been the backbone for several computer vision tasks including image recognition \cite{szegedy2015going, he2016deep, zoph2018learning}, object detection \cite{lin2017feature}, and image segmentation \cite{chen2017deeplab}. Their translation equivariance helps them generalize to different positions and spatial weight sharing helps reduce the trainable parameters. Despite these benefits, convolutional networks suffer from a few limitations. First, they are content agnostic in nature as the same weights are applied at all locations of an input feature map. Thus the content differences between pixels which could yield valuable downstream information are not taken into account. Second, both parameter count and floating-point operations (FLOPs) scale poorly with an increase in receptive field which is essential to capture long range interaction of pixels. To mitigate this, prior work has employed two methods. The first is knowledge distillation \cite{hinton2015distilling, zagoruyko2016paying,tian2019contrastive} wherein a more complex teacher model teaches a simpler student model through the transference of logits or features. The other approach is pruning \cite{Zhang_2018_ECCV, dettmers2019sparse, mostafa2019parameter, lee2018snip} which removes unnecessary weights which do not significantly affect the accuracy of the model. Recently, self-attention (SA) mechanisms have been used either alongside convolutions \cite{bello2019attention} or have completely replaced them in vision models \cite{parmar2019stand}, showing promising results for complex computer vision tasks. While self-attention lacks spatial context, positional embeddings can be used to make up for this limitation. Self-attention however, possesses a number of advantages. First, unlike convolutions, they consume fewer parameters and FLOPs, and scale much better with larger receptive fields enabling efficient capture of long range context without significantly increasing model complexity. They are also extremely parallelizable and have the potential to be accelerated in suitable hardware by exploiting parallel-execution \cite{park2020optimus}. In this paper, we propose a training framework that can take advantage of all three mechanisms to create extremely efficient vision models while also reducing the time it takes to optimize and fine tune them. Our contributions are as follows: We propose a joint model optimization framework called sparse distillation, wherein we exploit the benefits from pruning, distillation and self-attention mechanisms. Using a compute-heavy CNN with rich positional information as a teacher, we distill knowledge into a self-attention based student model. Simultaneously, we enforce the idea of sparse learning \cite{dettmers2019sparse, mostafa2019parameter, kundu2020tunable} of the student to yield a pruned self-attention model. Unlike many existing pruning schemes \cite{Zhang_2018_ECCV}, sparse distillation requires only a global target parameter density and only one pass of training, significantly reducing memory access and the computation burden associated with iterative optimization. We further extend our framework to support structured column pruning to yield models that can increase inference speeds as well offer the benefit of reduced parameters. We demonstrate the effectiveness of our proposed framework through extensive experiments on three popular datasets, CIFAR-10, CIFAR-100 \cite{krizhevsky2009learning} and Tiny-ImageNet \cite{hansen2015tiny}. When compared to unpruned baselines as well as convolutional counterparts, our models perform remarkably accurately while requiring a fraction of the parameters and FLOPs. The rest of the paper is organized as follows: Section \ref{sec:back} and \ref{sec:method} go over the details of our framework, Section \ref{sec:expt} details the results and Section \ref{sec:concl} presents our conclusions and planned future work.

\begin{figure*}[!t]
\includegraphics[width=0.97\textwidth]{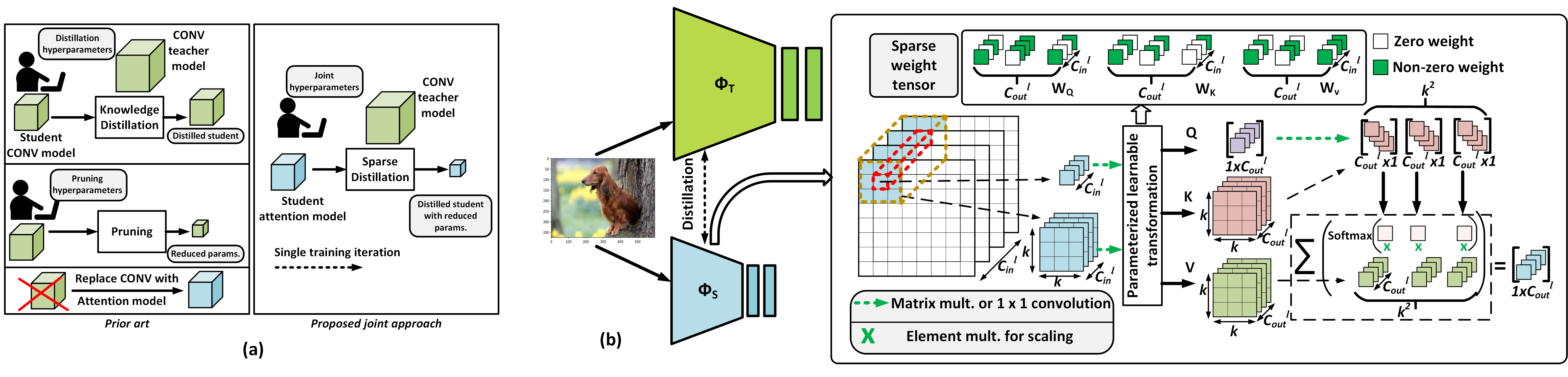}
\centering
   \caption{(a) Three orthogonal \textit{prior art} to generate reduced parameter models, namely distillation, pruning (through training) and use of self-attention models. $\textit{Proposed approach}$: A joint optimization strategy where distillation extracts the positional context and richer knowledge from the convolutional teacher, and pruning keeps the parameter budget of the student extremely limited. (b) Internal attention layer computation mechanics of the student $\Phi_S$, with only a fraction of the weights being non-zero. Note that we replace all $k \times k$ ($k$ $>$ $1$) convolutions with self-attention.}
\label{fig:sparse_distill_framework}
\vspace{-4mm}
\end{figure*}

\section{Preliminaries}
\label{sec:back}
\subsection{Self-attention}
\label{subsec:sa}

Given an input feature map at layer $l$ of a network, consider the pixel at position $(i,j)$, $x_{ij} \in \mathbb{R}^{C^l_{in}}$. A convolution operation centered at this pixel operates in a $k \times k$ neighborhood $\mathcal{N}_{k}(i,j)$ where $k$ is the kernel size. A single-headed local self-attention layer replacing this convolution, with a spatial extent $k$ considers all pixels from the neighbouring locations $(a,b)\in \mathcal{N}_{k}(i,j)$ and computes the output $y_{ij} \in \mathbb{R}^{C^l_{out}}$ as,
\begin{equation} \label{equation:standard-self-attention}
y_{ij} = \sum_{\mathclap{a, b \in \, \mathcal{N}_k(i, j)}}
        \texttt{softmax}_{a b}\left( \frac{q_{i j}^\top  k_{a b}}{\sqrt{C^l_{out}}}  + \frac{q_{i j}^\top r_{a-i,b-j}}{\sqrt[\leftroot{-2}\uproot{2}{4}]{C^l_{out}}} \right)  v_{a b}
\end{equation}
\noindent
Here $q_{ij}$ = $\mathbf{W_Q} x_{ij}$, $k_{ab}$ = $\mathbf{W_K} x_{ab}$, and $v_{ab}$ = $\mathbf{W_V} x_{ab}$ are the queries, keys, and values, respectively for pixel location ($i, j$). The term $r_{a-i,b-j}$ represents a simple relative positional embedding based on the offset $(a-i,b-j)$, and helps learn the spatial context. The major trainable parameters here are $\mathbf{W_Q}$, $\mathbf{W_K}$, $\mathbf{W_V} \in \mathbb{R}^{C^l_{in} \times C^l_{out}}$ and they do not increase with an increase in spatial extent $k$. We use multi-headed self-attention, wherein $N$ attention heads are each allowed to attend to $\frac{C^l_{out}}{N}$ output channels and following which the results of all the heads are concatenated to produce $y_{ij} \in \mathbb{R}^{C^l_{out}}$. 

We use the self-attention ResNet architectures introduced by Parmar \textit{et. al} \cite{parmar2019stand} (SA ResNet26 and SA ResNet38) as the student models in our work. We further use two variants of each model: the hybrid and homogeneous variants. The hybrid models use a convolutional block at the first layer of the architecture (also called the model's stem) and use self-attention to replace the remaining spatial-convolutions in the model. The homogeneous variant, on the other hand, uses self-attention layers throughout the model including the stem. In \cite{parmar2019stand}, the authors used a more sophisticated positional embedding while computing self-attention at the stem to compensate for the limited spatial awareness mentioned before. In our work, we uniformly use the same simple relative positional embedding as in equation \ref{equation:standard-self-attention} to promote consistency and ease of parallelism in hardware. 


\subsection{Knowledge Distillation and Model Pruning}
\label{subsec:kd}
We distill knowledge from both logits and feature maps to improve the performance of our student models. We use traditional softmax temperature based knowledge distillation \cite{hinton2015distilling} for transferring knowledge from logits and employ attention transfer \cite{zagoruyko2016paying} to transfer knowledge from the teacher's activation maps. This enables the student to learn both spatial context as well as the distribution of class probabilities while being pruned during training. 

Pruning achieves inference parameter reduction by removing unimportant weights from a model, and can be broadly classified into two categories, irregular pruning and structured pruning. The former prunes weight scalars based on their importance and enjoys the advantage of lower parameter density ($d$) with similar accuracy as an unpruned model. The same cannot always be said for structured pruning that prunes at the granularity of filters, channels or columns. However, irregularly pruned models suffer from the overhead of non-zero weight indices \cite{kundu2020pre} and often requires dedicated hardware to extract compression and speedup benefits. Despite the larger parameter density requirement, structured pruning can yield inference speedup without dedicated hardware support \cite{liu2018rethinking}. Our sparse distillation framework, detailed next, exploits the benefits from both distillation and pruning and supports both irregular and structured pruning. 

\section{Sparse Distillation}
\label{sec:method}
Given a layer $l$ with activation tensor $\textbf{A}^l \in \mathbb{R}^{H_{in}^l \times W_{in}^l \times C_{in}^l}$ we define the activation-based mapping function, $\mathcal{F}^p$ = $\sum_{c=1}^{C_{in}^l}|\textbf{A}_c|^p$, where $p \geq 1$. Here, $C_{in}^l$ represents the channels of spatial dimensions $H_{in}^l \times W_{in}^l$ and $\mathcal{F}^p$ is the flattened spatial attention map \cite{zagoruyko2016paying}. We denote the teacher and student models as $\Phi_T$ and $\Phi_S$, respectively. Let $\Psi_S^m$ and $\Psi_T^m$ represent the $m^{th}$ pair of vectorized attention maps $\mathcal{F}$ of specific layers of $\Phi_S$ and $\Phi_T$, respectively. Our proposed loss function can be defined as:
\begin{align}\label{eq:sp_distill_loss}
        {\mathcal{L}}_{total} = \alpha{\mathcal{L}}_{S}(y,y^S) + & (1-\alpha){\mathcal{L}}_{KL}\left(\sigma\left(\frac{z^T}{\rho}\right),\sigma\left(\frac{z^S}{\rho}\right)\right)\\ \nonumber
    + & \frac{\beta}{2}\sum_{m \in I}\left\lVert\frac{\Psi_S^m}{\lVert \Psi_S^m\rVert_2} - \frac{\Psi_T^m}{\lVert \Psi_T^m\rVert_2}\right\lVert_2 ,
\end{align}
\noindent
where the first term, $\mathcal{L}_{S}$, corresponds to the cross entropy loss of the self-attention student obtained by comparing the true logits ($y$) and the predicted logits ($y^S$). The second term, $\mathcal{L}_{KL}$, represents the KL-divergence loss (KD-loss) between the teacher ($\Phi_T$) and the student ($\Phi_S$) transferring knowledge via logits. $\sigma$ represents the softmax function with $\rho$ being its temperature.  The last term defines the activation-based attention transfer loss (AT-loss) between the two. As proposed in \cite{zagoruyko2016paying} we use the $l_2$-norm of the normalized attention-maps to compute the loss. The parameters $\alpha$ and $\beta$ control the influence of each distillation method.

\begin{algorithm}[t]
\footnotesize
\SetAlgoLined
\DontPrintSemicolon
\textbf{Input}: \textit{totalEpochs},  momentum contribution $\pmb{\mu}^l$, prune rate $p$ ($p_{e=0}$), initial $\mathbf{W}$, initial mask $\mathbf{\Pi}$, target parameter density $d$, teacher model $\Phi_T$, student model $\Phi_S$.\\
\textbf{Output}: Sparse distilled $\Phi_S$.\\
\For{$l \leftarrow 0$ \KwTo $L$}
{
    $\text{// Initialize weights } {\mathbf{W}^l} \text{, mask } {\mathbf{\Pi}^l} \text{, and apply mask } {\mathbf{\Pi}^l} $\;
}
\For{$\text{e} \leftarrow 0$ \KwTo \text{totalEpochs}}
{
   
    \For{$\text{t} \leftarrow 0$ \KwTo \text{totalBatches}}
    {
     $\text{// Evaluate student loss } {\mathcal{L}_{total}} \text{ and the gradient} $\;
     $\text{// Update weights and momentum contribution}$\;
     \For{$l \leftarrow 0$ \KwTo $L$}
     {
     $\text{// Apply mask to weights}$\;
     }
    }
    $\text{// Evaluate total momentum}$\;
    $\text{// Get total weights to be pruned}$\;
    $\text{// Linearly decay prune rate } p_e$\;
    \For{$l \leftarrow 0$ \KwTo $L$}
    {
        $\text{// Update layer momentum contribution } \pmb{\mu}^l$\;
        $\text{// Prune fixed \% of active weights from each layer}$\;
        $\text{// Regrow fraction of inactive weights based on } \pmb{\mu}^l$\;
        $\text{// Update mask for next epoch}$\;
  }
}
 \caption{Sparse distillation}
 \label{alg:spd}
\end{algorithm}

To prune the student model while simultaneously distilling knowledge from the teacher, we first update $\Phi_S$'s total trainable parameters and then use the mask to forcibly set a fraction of these parameters to zero. Inspired by the idea of sparse-learning \cite{dettmers2019sparse, kundu2020tunable}, we start the distillation with initialized weights and a random pruning mask that satisfies the non-zero parameter budget corresponding to the target parameter density $d$ for $\Phi_S$. Based on the loss from equation \ref{eq:sp_distill_loss}, we evaluate the layer’s importance by computing the normalized momentum contributed by its non-zero weights during an epoch. This enables us to decide which layers should have more non-zero weights under the given parameter budget and we update the pruning mask accordingly. Concretely, we re-grow the weights with the highest momentum magnitude after pruning a fixed percentage of the least-significant weights from each layer based on their magnitude \cite{dettmers2019sparse}. Details of the sparse distillation training is presented in Algorithm \ref{alg:spd}.

To reduce the effective model size and potentially speed up inference \cite{liu2018rethinking}, our framework also supports column pruning, a form of structured pruning. Let the weight tensor of a convolutional layer $l$ be denoted as $\mathbf{W}^l \in \mathbb{R}^{C^l_{out} \times C^l_{in} \times k \times k}$, where $C^l_{out}$ and $C^l_{in}$ represents number of filters and channel per filter, respectively, and $k$ represent filter size. We convert this tensor to a 2D weight matrix  with $C^l_{out}$ rows and $C^l_{in} \times k \times k$ columns. Next, we partition this matrix into $C^l_{in} \times k \times k$ sub-matrices of $C^l_{out}$ rows and $1$ column. To compute the importance of a column representing the $(k_h,k_w)^{th}$ entry of $c^{th}$ channel of the filters, we find the Frobenius norm (F-norm) of corresponding sub-matrix, thus effectively compute $O^l_{c,k_h,k_w}$ = $||{\mathbf{W}^l_{:,c,k_h,k_w}}||^{2}_F$. Based on the fraction of non-zero weights that need to be regrown during an epoch $e$, $p_e$, we compute the number of columns that must be pruned from each layer, ${c}_{p_e}^l$, and prune the ${c}_{p_e}^l$ columns with the lowest F-norms.  Then based on the layer importance measure (through momentum) we determine the number of zero-F-norm columns $r^l_e \geq 0$ that should be re-grown for each layer $l$. Thus, we re-grow the $r^l_e$ zero-F-norm columns with the highest F-norms of their momentum.

\begin{figure*}
  \begin{subfigure}[b]{0.36\columnwidth}
    \includegraphics[width=\linewidth]{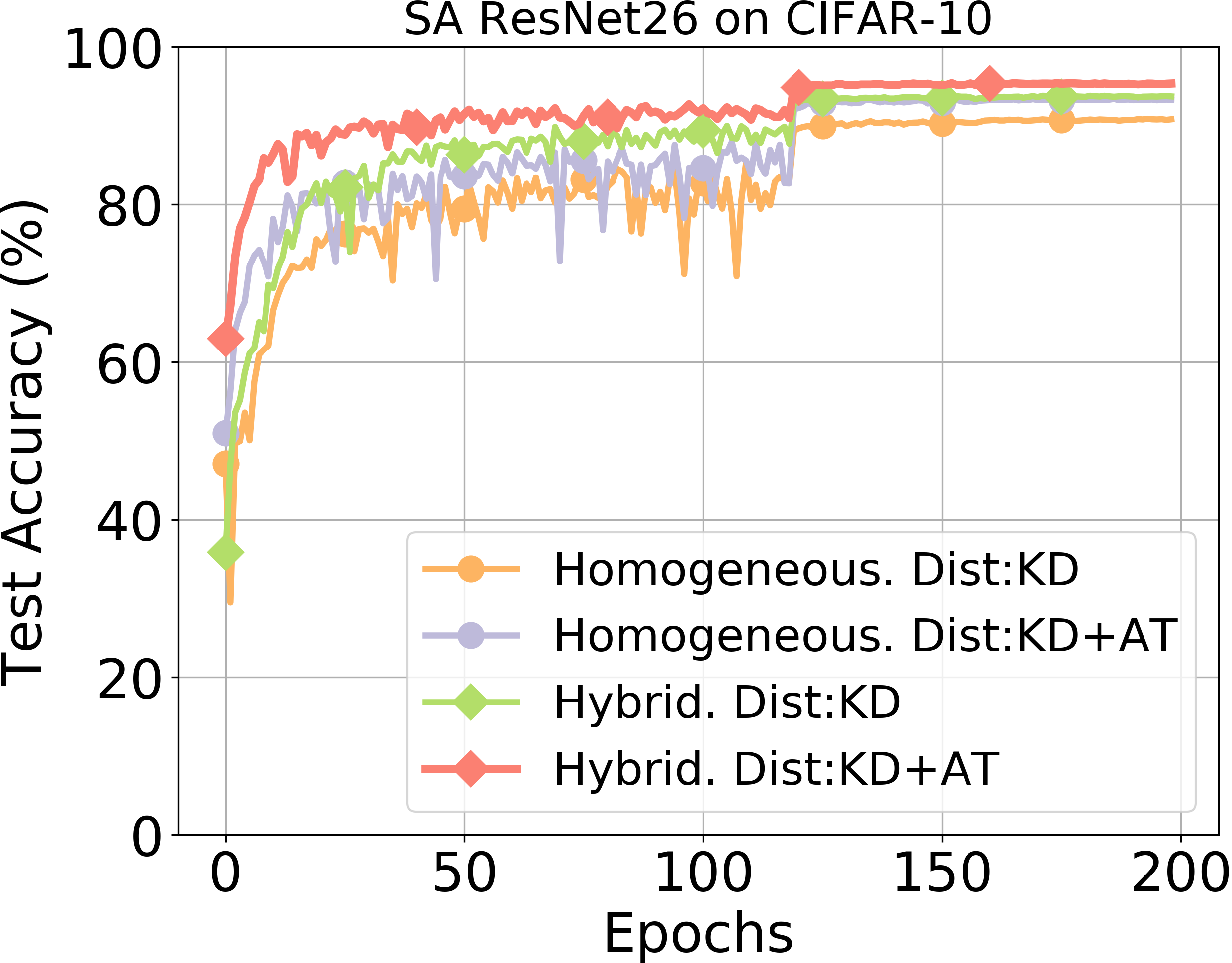}
    \caption{}
    \label{fig:1a}
  \end{subfigure}
  \hfill 
  \begin{subfigure}[b]{0.36\columnwidth}
  \includegraphics[width=\linewidth]{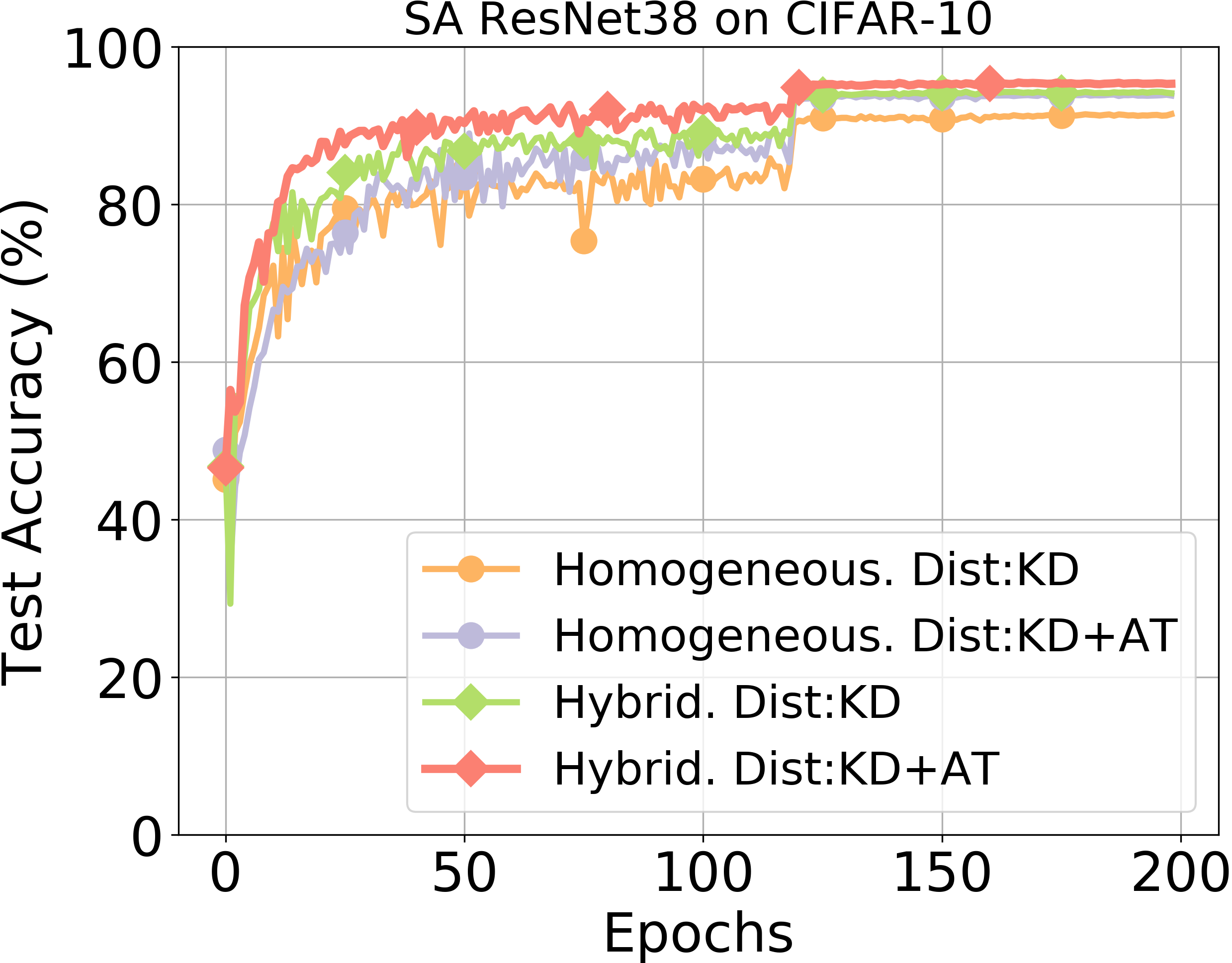}
    \caption{}
    \label{fig:1b}
  \end{subfigure}
  \hfill 
  \begin{subfigure}[b]{0.36\columnwidth}
      \includegraphics[width=\linewidth]{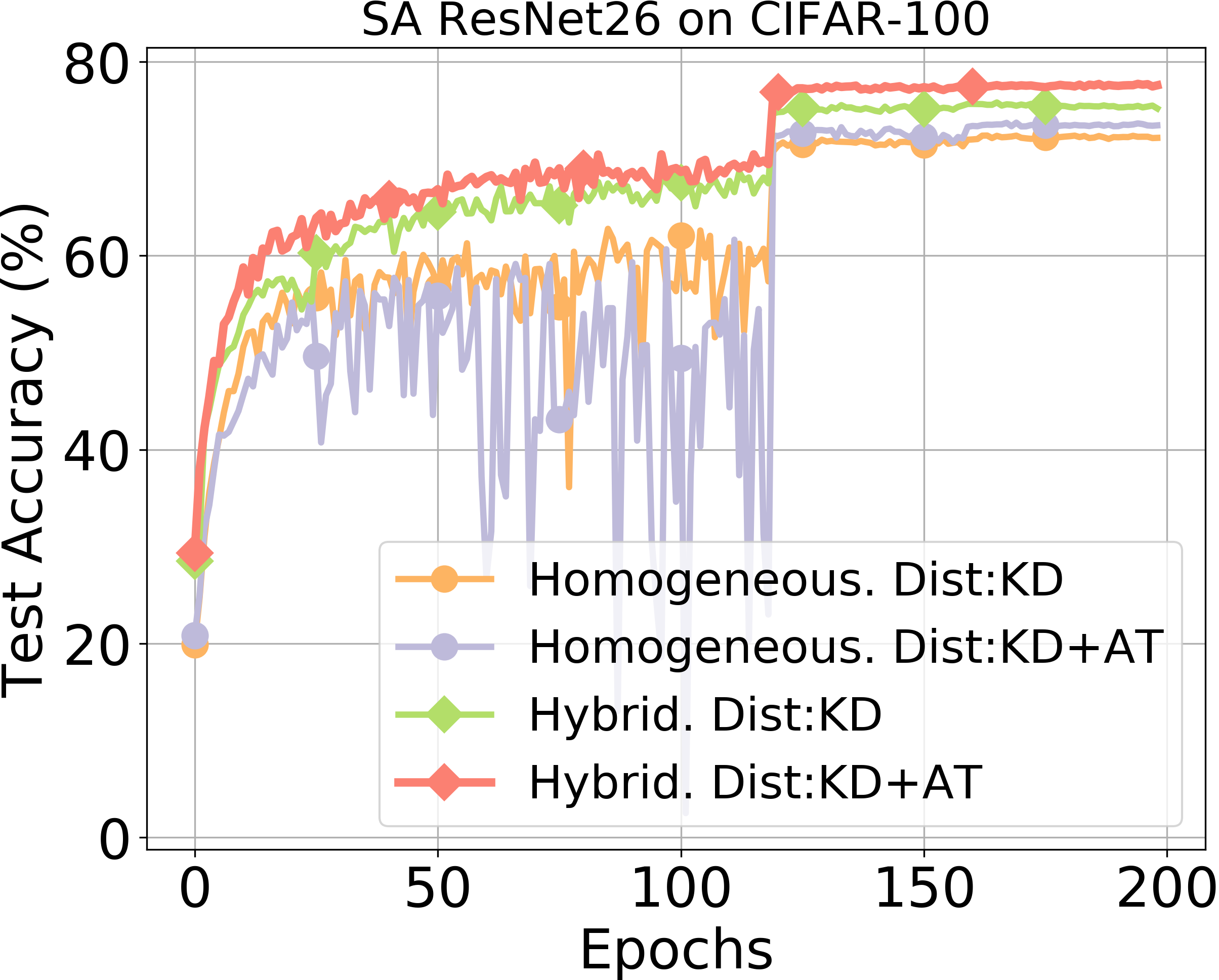}
    \caption{}
    \label{fig:1c}
  \end{subfigure}
  \hfill 
  \begin{subfigure}[b]{0.36\columnwidth}
      \includegraphics[width=\linewidth]{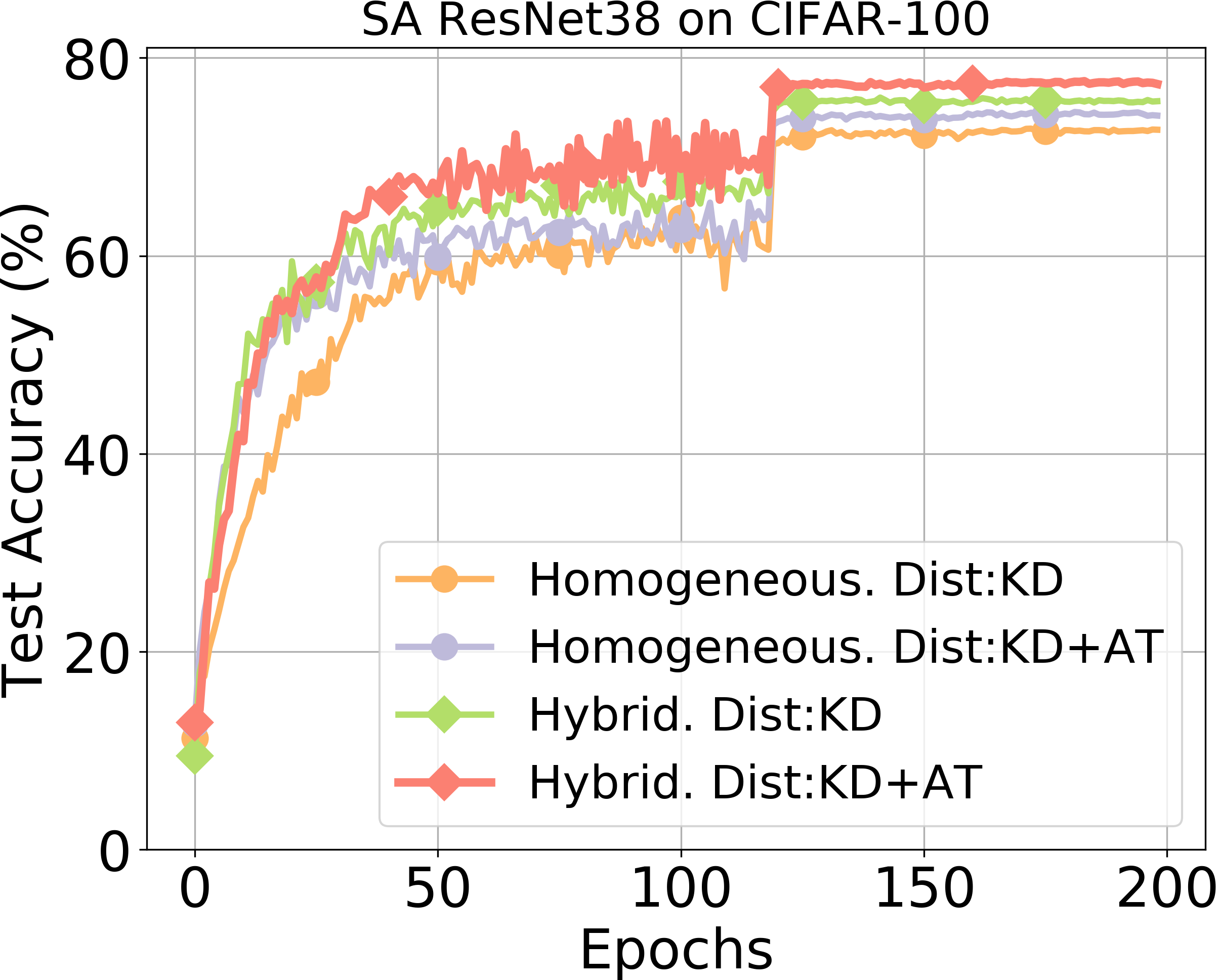}
    \caption{}
    \label{fig:1d}
  \end{subfigure}
  \hfill 
  \begin{subfigure}[b]{0.36\columnwidth}
      \includegraphics[width=\linewidth]{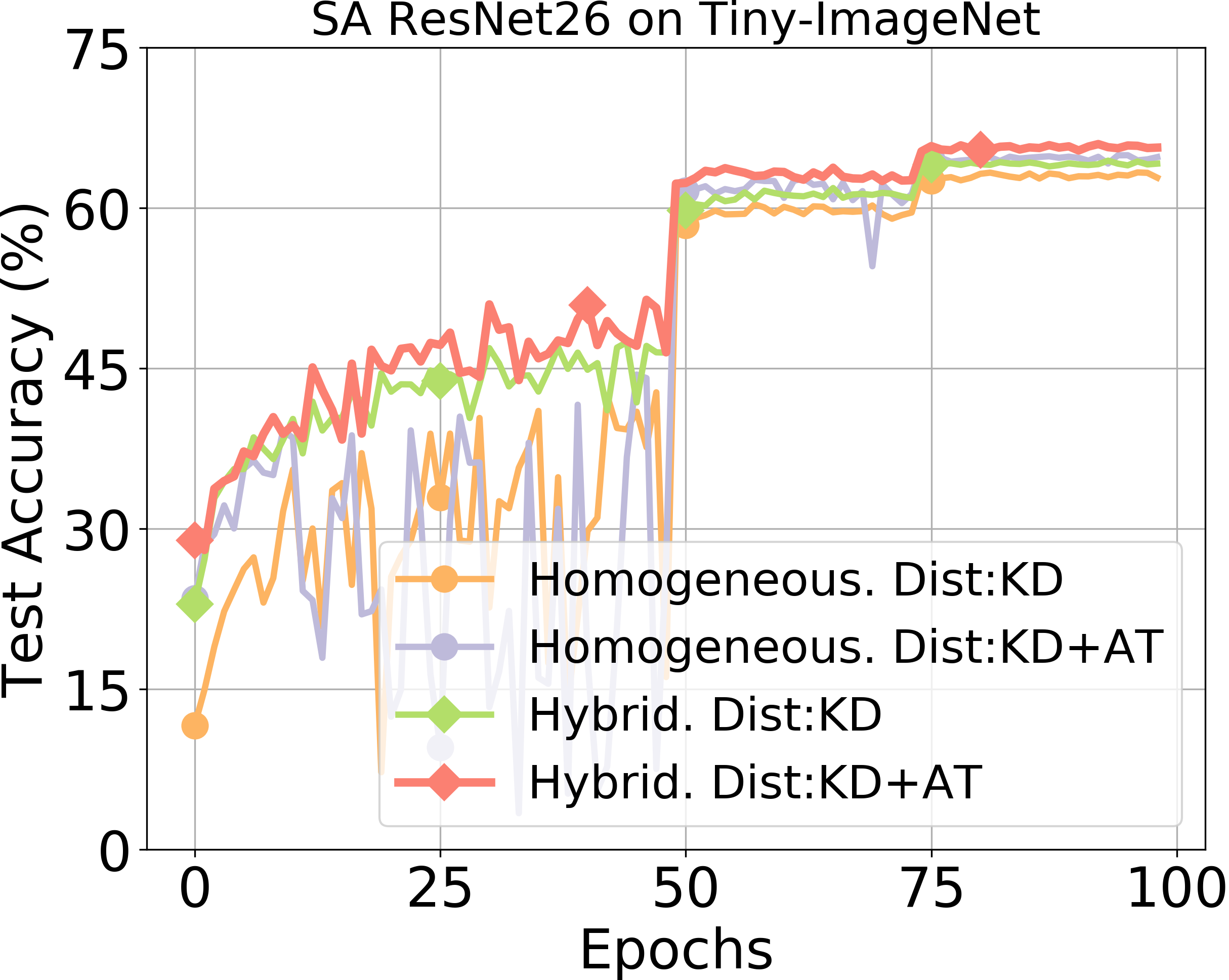}
    \caption{}
    \label{fig:1e}
  \end{subfigure}
  \caption{Test accuracy vs. epochs plots for the AttentionLite models with a target parameter density $d$ = $0.25$ (with irregular pruning) generated through sparse distillation framework.}
  \label{fig:test_acc_vs_epoch}
  \vspace{-2mm}
\end{figure*}

\begin{figure*}
  \begin{subfigure}[b]{0.36\columnwidth}
    \includegraphics[width=\linewidth]{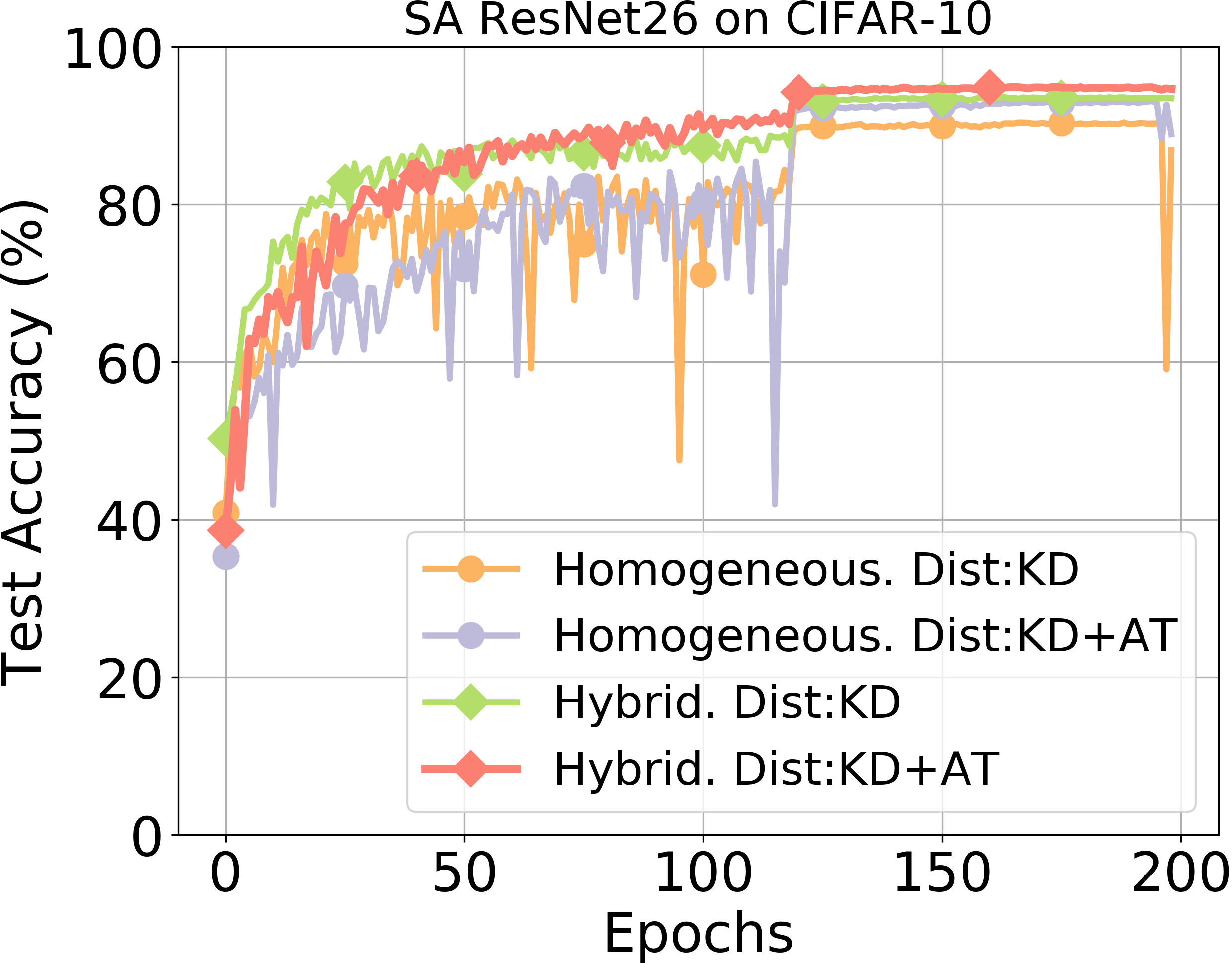}
    \caption{}
    \label{fig:col_a}
  \end{subfigure}
  \hfill 
  \begin{subfigure}[b]{0.36\columnwidth}
  \includegraphics[width=\linewidth]{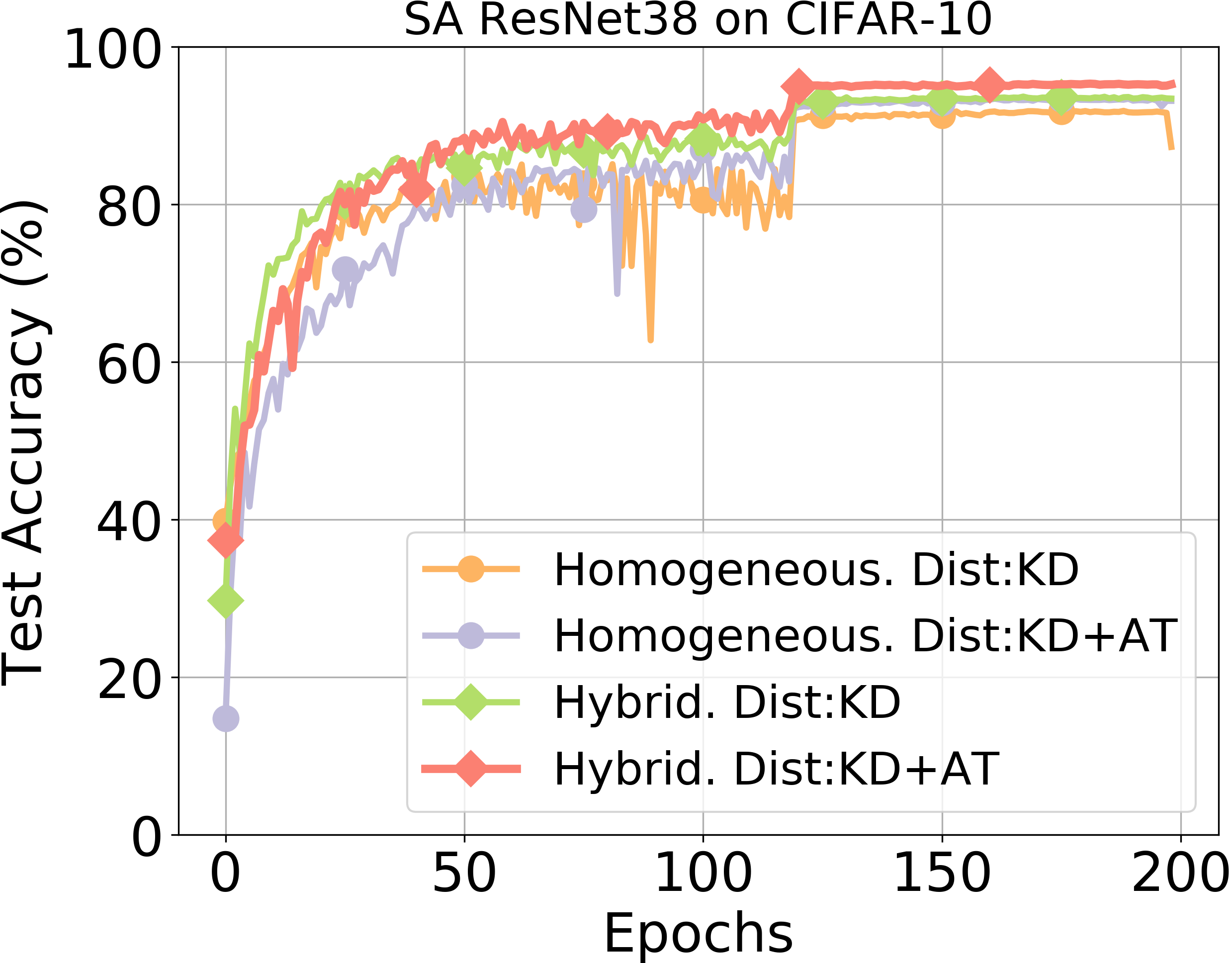}
    \caption{}
    \label{fig:col_b}
  \end{subfigure}
  \hfill 
  \begin{subfigure}[b]{0.36\columnwidth}
      \includegraphics[width=\linewidth]{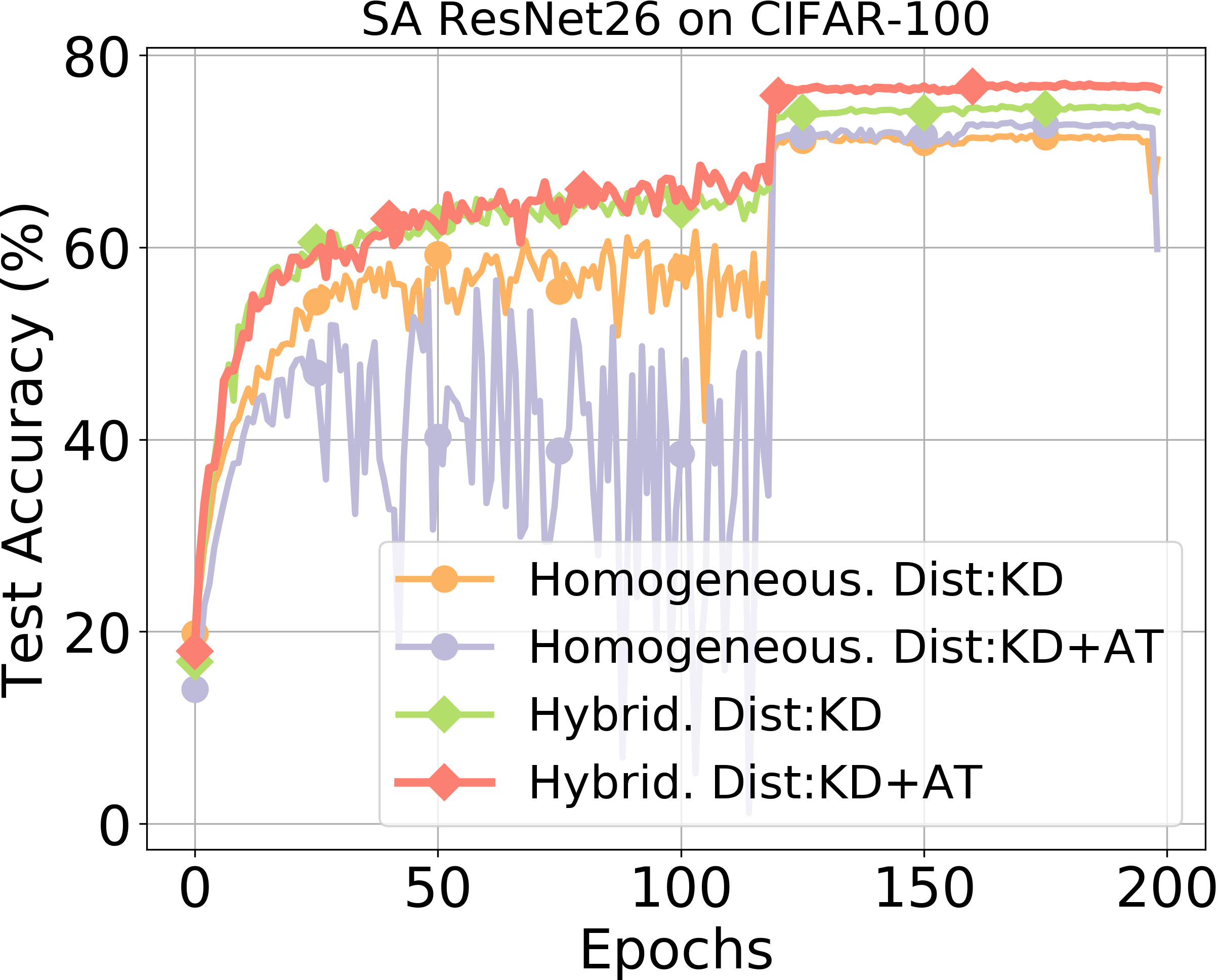}
    \caption{}
    \label{fig:col_c}
  \end{subfigure}
  \hfill 
  \begin{subfigure}[b]{0.36\columnwidth}
      \includegraphics[width=\linewidth]{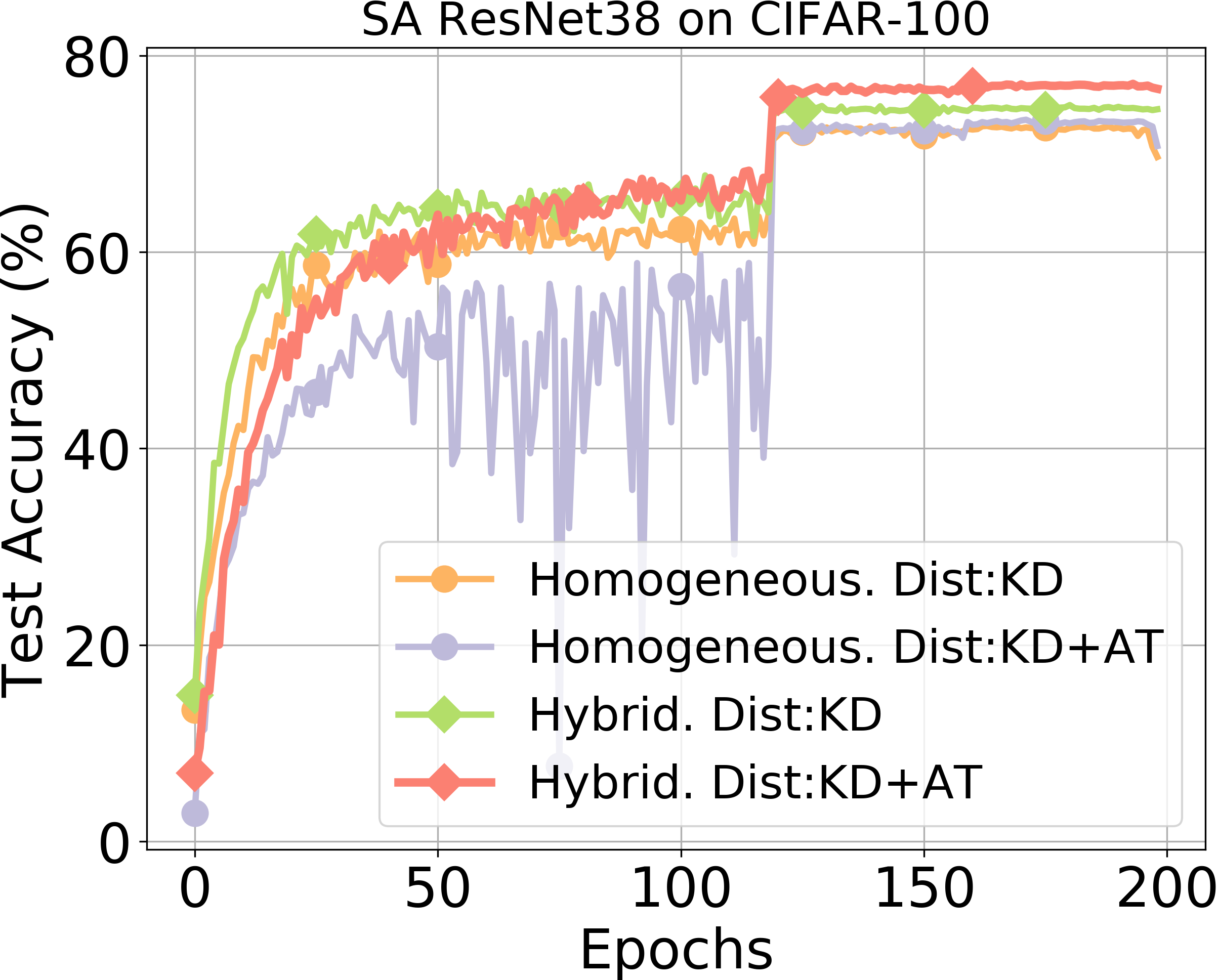}
    \caption{}
    \label{fig:col_d}
  \end{subfigure}
  \hfill 
  \begin{subfigure}[b]{0.36\columnwidth}
      \includegraphics[width=\linewidth]{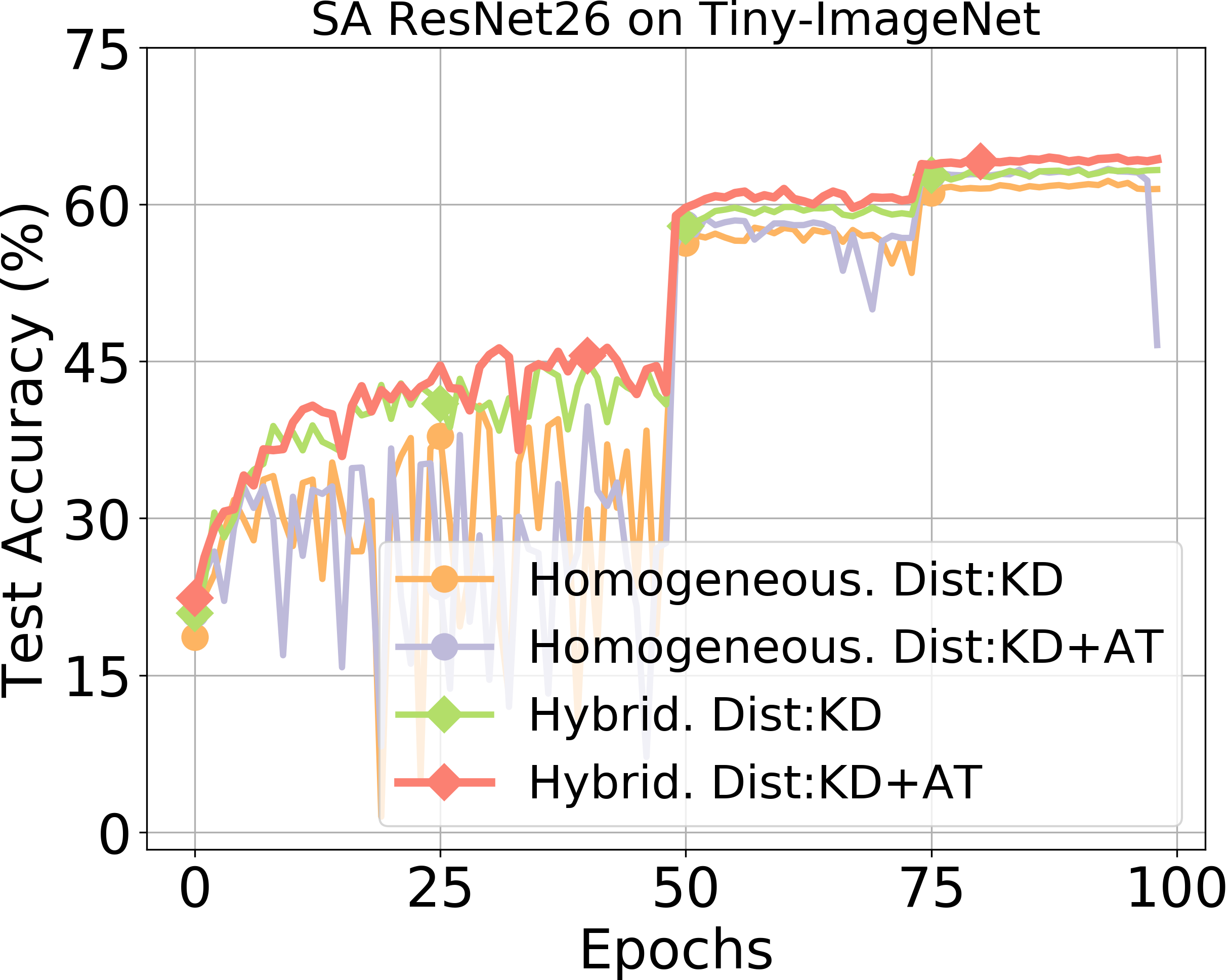}
    \caption{}
    \label{fig:col_e}
  \end{subfigure}
  \caption{Test accuracy vs. epochs plots for the AttentionLite models with a target parameter density $d$ = $0.5$ (with column pruning) generated through sparse distillation framework.}
  \label{fig:test_acc_vs_epoch_column}
  \vspace{-3mm}
\end{figure*}

\begin{table}[!t]
\begin{center}
\scriptsize\addtolength{\tabcolsep}{-5.5pt}
\begin{tabular}{|c|c|c|c|c|c|c|c|c|}
\hline
Hybrid & Dataset & Distill. & Pruning & Accuracy  &\multicolumn{2}{|c|}{Baseline acc. (\%)} & Param.  & FLOPs \\
\cline{6-7}
SA &  {} & type & type  & (\%) & Teacher & Student & reduction & reduction \\
student &  {} &  &   &  & &  &  &  \\
\hline
\hline
SA       & CIFAR & KD     & Irregular & 93.28 & 95.34 & 92.77 & $29.75\times$ & $1.96\times$  \\
ResNet26 & -10   &        & Column    & 93.74 &  & & $5.81\times$ & $\textbf{2.02}\times$  \\
\cline{3-5} \cline{8-9}
{}  &       &  KD+AT & Irregular & 94.7  &  &  & $\textbf{30.26}\times$ & $1.91\times$  \\
{}       &       &        & Column    & \textbf{94.88}  &  & & $5.83\times$ & $\textbf{2.02}\times$   \\
\cline{2-9}
{}       & CIFAR & KD     & Irregular & 74.48 & 79.04 & 72.33 &  $28.20\times$ & $1.89\times$ \\
{} & -100   &        & Column    & 74.68  &   &  & $5.64\times$ & $\textbf{2.02}\times$  \\
\cline{3-5} \cline{8-9}
{}  &       &  KD+AT & Irregular & 76.31  &  & & $\textbf{28.59}\times$ & $1.89\times$  \\
{}       &       &        & Column    & \textbf{77.01}  &  &  & $5.70\times$ & $\textbf{2.02}\times$  \\
\cline{2-9}
{}       & Tiny- & KD     & Irregular & 64.40 & 67.10 & 63.72 & $\textbf{11}\times$ & $1.97\times$   \\
{} & ImageNet   &        & Column    & 63.71  &  &  & $5.58\times$ & $\textbf{2.11}\times$  \\
\cline{3-5} \cline{8-9}
{}  &       &  KD+AT & Irregular & \textbf{65.88}  &  &  & $\textbf{11}\times$ & $1.97\times$  \\
{}       &       &        & Column    & 64.42  &  &  & $5.50\times$ & $\textbf{2.11}\times$  \\
\hline
SA       & CIFAR & KD     & Irregular & 93.82 & 95.34 & 92.72 & $20.15\times$ & $\textbf{1.61}\times$  \\
ResNet38 & -10   &        & Column    & 93.89 &  &  & $3.97 \times$ & $1.55\times$  \\
\cline{3-5} \cline{8-9}
{}  &       &  KD+AT & Irregular & 95.03  &  &  & $\textbf{20.34}\times$ & $1.52\times$  \\
{}       &       &        & Column    & \textbf{95.20}  &  &  & $3.97 \times$ & $1.56\times$  \\
\cline{2-9}
{}       & CIFAR & KD     & Irregular & 74.97 & 79.04 &  72.49 & $\textbf{19.92}\times$ & $1.50\times$  \\
{} & -100   &        & Column    & 74.55  &   &  & $3.93\times$ & $1.55\times$   \\
\cline{3-5} \cline{8-9}
{}  &       &  KD+AT & Irregular & 76.59  &  &  &  $19.75\times$ & $1.48\times$  \\
{}       &       &        & Column    & \textbf{76.93}  &  &  &  $3.89\times$ & $\textbf{1.57}\times$  \\
\hline
\end{tabular}
\end{center}
\caption{Performance of hybrid variants optimized by sparse distillation. We use $d$ = 0.1 for the CIFAR datasets, and $d$ = 0.25 for Tiny-ImageNet with irregular pruning. For channel pruning we use $d$ = 0.5 for all datasets.}
\label{tab:hybrid_sa_perfm}
\vspace{-4mm}
\end{table}

\vspace{-3mm}
\section{Experimental Results}
\label{sec:expt}
In this section, we describe our experimental setup and demonstrate that AttentionLite models achieve a very favourable trade-off between efficiency and accuracy when compared to their unoptimized counterparts as well as their compute-heavy teacher. We present the results of evaluating our models on three widely used datasets, namely, CIFAR-10 \cite{krizhevsky2009learning}, CIFAR-100 \cite{krizhevsky2009learning} and Tiny-ImageNet \cite{hansen2015tiny} and share our insights from these results below. 

\subsection{Experimental Setup} 
\label{subsec:setup}
For all three datasets, we perform the horizontal flip and random crop with reflective padding as augmentations. We trained the models for a total of 200 epochs with a batch size of 100 on the CIFAR datasets and for 100 epochs with a batch size of 32 on Tiny-ImageNet. An initial learning rate of $0.1$ with a weight decay of $10^{-4}$ was used for all three datasets. We reduced the learning rate by $0.1$ after epochs 120, 160 and 180 for the CIFAR datasets, while for Tiny-ImageNet, we reduced the learning rate by the same factor after half and three-quarters of the training had elapsed. As per Tian \textit{et. al} \cite{tian2019contrastive}, we set $\alpha$ = $0.1$ and $\beta$ = $1000$ for training the self-attention student and applied attention transfer between the models after every residual block. For the self-attention layers we used 8 heads and used ResNet50 as the convolutional teacher for all experiments.


\subsection{Discussion of AttentionLite Model Performance} 
\label{subsec:res_hybrid_lsa}

\begin{table}
\begin{center}
\scriptsize\addtolength{\tabcolsep}{-5pt}
\begin{tabular}{|c|c|c|c|c|c|c|c|c|}
\hline
Homog- & Dataset  & Pruning & Accuracy &\multicolumn{3}{|c|}{Baseline acc. (\%)}  & Param. &  FLOPs \\
\cline{5-7}
-eneous      & {}       & type   & (\%)    &  Teacher & Hybrid      &  Homog-  & reduction      & reduction        \\
SA &  {} &    &  &  & student & -eneous  & &  \\
 student  &     &    &  &  &         & student  & & \\
\hline
\hline
SA       & CIFAR &    Irregular & $\textbf{93.34}$ & 95.34 & 92.77 & 88.74 & $\textbf{11.51}\times$ & $1.87\times$  \\
ResNet26 & -10   &    Column    & 93.05 &  & &  & $5.75 \times$ & $\textbf{2.02}\times$  \\
\cline{2-9}
{}       & CIFAR &   Irregular & \textbf{73.71} & 79.04 & 72.33 & 65.8 & $\textbf{11.51}\times$ & $1.86\times$  \\
{} & -100   &    Column    & 72.98  & & &  & $5.72\times$ & $\textbf{2.01}\times$ \\
\cline{2-9}
{}       & Tiny- &  Irregular & \textbf{64.92} & 67.1 & 63.72 & 61.1 & $\textbf{11}\times$ & $1.97\times$  \\
{} & ImageNet   &   Column    & 63.39 & & &  & $5.50 \times$ & $\textbf{2.11}\times$  \\
\hline
SA       & CIFAR &  Irregular & \textbf{93.93} & 95.34 & 92.72 & 90.78 &  $\textbf{8}\times$ & $1.46\times$  \\
ResNet38 & -10   &   Column    & 93.39 & & &  & $4.01 \times$ & $\textbf{1.56}\times$  \\
\cline{2-9}
{}       & CIFAR &  Irregular & \textbf{74.53} & 79.04 & 72.49 & 67.73 & $\textbf{7.87}\times$ & $1.45\times$  \\
{} & -100   &   Column    & 73.46  &  & &  & $3.95\times$ & $\textbf{1.56}\times$  \\
\hline
\end{tabular}
\end{center}
\caption{Performance of homogeneous variants optimized by sparse distillation (with KD+AT loss). Here $d$ = 0.25 for irregular pruning and 0.5 for channel pruning. For brevity, we omit the results with just KD loss. Those results can be found in \ref{fig:test_acc_vs_epoch} and \ref{fig:test_acc_vs_epoch_column}.}
\label{tab:homo_sa_perfm}
\vspace{-3mm}
\end{table}

The performance of the hybrid variants can be seen in table \ref{tab:hybrid_sa_perfm}. The baseline accuracy column shows the performance of the teacher as well as the unoptimized student model as comparison points. Since the intended goal is to replace the teacher with the AttentionLite models, columns 8 and 9 represent parameter and FLOPs reduction achieved by the these models compared to their teacher. Table \ref{tab:homo_sa_perfm} details the performance of the homogeneous variants.

 From these results, we can derive the following insights. First, with the hybrid variants, our sparse distillation framework can yield parameter reduction of up to $30.26 \times$, $28.59 \times$, and $11 \times$ on CIFAR-10, CIFAR-100, and Tiny-ImageNet respectively with irregular pruning. However, column pruning yields a FLOPs reduction of up to $2.02 \times$, $2.02 \times$, and $2.11 \times$ on the same datasets. As column pruning has stricter pruning constraints compared to irregular pruning, it achieves a smaller parameter reduction. With both forms of pruning, we can see that AttentionLite models incur a tiny drop in accuracy compared to their teacher but outperform their unoptimized counterparts significantly. The homogeneous variants achieve a parameter reduction of up to $11.51\times$ on CIFAR-10 and CIFAR-100, and $11\times$ on Tiny ImageNet. They have a FLOPs reduction of $2.02\times$, $2.01\times$, and $2.11\times$, respectively, when compared to their teacher. The homogeneous models suffer accuracy drops when pruned aggressively and are pruned less (for irregular pruning) in our experiments to preserve accuracy. Hence, they achieve lesser parameter reduction. However, these models achieve good results just with simple structural blocks throughout without any complex positional embeddings. Thus, homogeneous variants have the potential to increase parallel execution throughput with no additional architectural overhead. Due to the absence of overlapping patch-based compute, homogeneous self-attention models can avoid compute-inefficient splitting of inputs, which is necessary for traditional CNNs to classify high-resolution images \cite{hou2019high}. We also observe from figures \ref{fig:test_acc_vs_epoch} and \ref{fig:test_acc_vs_epoch_column} that transferring logits as well as features via distillation (KD + AT) consistently yields better accuracy (up to $2.33\%$) across different models, pruning types and datasets. This seems to suggest that the Attention transfer loss can compensate for some of the lack of spatial context in the local self-attention layers.

\vspace{-2mm}
\section{Conclusions}
\label{sec:concl}
In this paper, we presented a novel joint optimization framework which produces parameter and compute efficient models in a single pass of training. Our experiments show that it is possible to produce models that can perform remarkably well compared to complex compute heavy CNNs while consuming only a fraction of the parameters and FLOPs. Hybrid AttentionLite models offer better accuracy while homogeneous variants offer the advantage of parallel execution. Our framework is complementary to existing schemes for producing efficient models and can work well with convolution based students as well. There are a number of potential directions to explore with our framework. While we have shown the efficacy of our approach on image classification, there are other tasks such as object detection, image segmentation and captioning which could benefit from our approach. Given the recent research in using Transformer-based architectures for vision tasks, sparse distilling a Transformer model to produce an efficient yet accurate network is an interesting direction. Evaluating the gains obtained by benchmarking the parallelism friendly AttentionLite models on custom hardware can inspire the industry to move towards optimized hardware support for attention mechanisms. 



\vfill\pagebreak

\bibliographystyle{IEEEbib}
\bibliography{strings,refs}

\end{document}